\documentclass[letterpaper, 10 pt, conference]{ieeeconf}  

\IEEEoverridecommandlockouts                              

\usepackage{cite}
\usepackage{bm}
\usepackage{amsmath}
\usepackage{amssymb}
\usepackage{graphics} 
\usepackage{epstopdf}
\usepackage{epsfig}

\newcommand{\mat}[1]{\mathbf{#1}}
\makeatletter
\newcommand{\vv}[1]{{\mathbf{\it #1}}\@ifnextchar_{\hspace{-0.3ex}}{}}
\newcommand{\vvc}[1]{{\bm{#1}}}
\makeatother

\newcommand{\gbf}[1]{{\bm{#1}}}
\newcommand{\gbfl}[1]{{\bm{#1}}}

\newcommand{\tq}{\gbfl{\tau}}
\newcommand{\J}{{\mat{J}}}
\newcommand{\M}{{\mat{M}}}
\newcommand{\Sm}{{\mat{S}}}
\newcommand{\q}{{\bm{q}}}
\newcommand{\qd}{{\bm{\dot q}}}
\newcommand{\qdd}{{\bm{\ddot q}}}
\newcommand{\xdd}{{\bm{\ddot x}}}

\newcommand{\h}{\mathbf{h}}
\newcommand{\tr}{\gbfl{\tau}}

\title{\LARGE \bf
Learning Task-Specific Dynamics to Improve Whole-Body Control
}

\author{Andrej Gams$^{1}$, Sean A. Mason$^{2}$, Ale\v{s} Ude$^{1}$, Stefan Schaal$^{2}$ and Ludovic Righetti$^{3,4}$
\thanks{*This work was supported by the Slovenian Research Agency project BI-US/16-17-063, New York University, the Max-Planck Society and the European Union’s Horizon 2020 research and innovation programme (grant agreement No 780684 and European Research Council’s grant No 637935).}
\thanks{$^{1}$Humanoid and Cognitive Robotics Lab, Dept. of Automatics, Biocybernetics and Robotics, Jo\v{z}ef Stefan Institute, Ljubljana, Slovenia.
        {\tt\small name.surname@ijs.si}}%
\thanks{$^{2}$Computational Learning and Motor Control Lab, University of Southern California, Los Angeles, California, USA. {\tt\small name.surname@usc.edu}}%
\thanks{$^{3}$Tandon School of Engineering, New York University, New York, USA. {\tt\small ludovic.righetti@nyu.edu}}%
\thanks{$^{4}$Max  Planck  Institute  for  Intelligent
Systems, Tuebingen, Germany.}%
}

\begin{document}

\maketitle
\thispagestyle{empty}
\pagestyle{empty}

\begin{abstract}
In task-based inverse dynamics control, reference accelerations used to follow a desired plan can be broken down into feedforward and feedback trajectories. The feedback term accounts for tracking errors that are caused from inaccurate dynamic models or external disturbances. On underactuated, free-floating robots, such as humanoids, good tracking accuracy often necessitates high feedback gains,
which leads to undesirable stiff behaviors. The magnitude
of these gains is anyways often strongly limited by the control bandwidth.
In this paper, we show how to reduce the required contribution of the feedback controller by incorporating learned task-space reference accelerations. Thus, we i) improve the execution of the given specific task, and ii) offer the means to reduce feedback gains, providing for greater compliance of the system. 
In contrast to learning task-specific joint-torques, which might produce a similar effect but can lead to poor generalization, our approach directly learns the task-space dynamics of the center of mass of a humanoid robot.
Simulated and real-world results on the lower part of the Sarcos Hermes humanoid robot demonstrate the applicability of the approach.

\end{abstract}

\section{INTRODUCTION} \label{sec:Intro}
Unmodeled dynamics (i.e. friction, link flexibilities, unmodeled actuator dynamics or approximate model parameters) can have severe effects on tracking performance of legged robots, and can be problematic not only for balancing but also to properly achieve other tasks. 
Models, however, are often difficult to obtain and/or incorrect, and while parameter identification can improve their quality \cite{Mistry2009,Wensing:2017bd}, it does not take into account unmodeled dynamic effects. The lack of accurate models has led to the use of combined feedforward and feedback control, which preserves stability and robustness to disturbances. The greater the error of the model, the greater the feedback gains necessary to ensure robust task achievement. This comes at the cost of a significant increase in stiffness and damping requirements, which can be a problem due to limited control bandwidth and which is often undesirable to prevent high impedance behaviors.

Different methods of acquiring dynamic models \cite{Nguyen-Tuong2011} and exploiting them in control have been proposed \cite{Franklin2011}. Optimization based approaches, such as hierarchical inverse dynamics \cite{Herzog2016,escande-ijrr-14,Kuindersma:2013tz}, have gained in popularity in the recent years for the control of legged robots. However, these approaches rely on dynamics models and often necessitate high task-space feedback gains to ensure good tracking performance on real robots which do not have accurate dynamic models.

Acquiring dynamic models of robots and tasks can be partially offset by iterative learning of the control signals, which relies on one of the main characteristics of robots: repeatability of the control actions from the same input signals. Iterative learning control (ILC) \cite{Bristow2006} was extensively applied in robotics, including for learning task-specific joint control torques \cite{Denisa2016}.
\begin{figure}[t!]
   \vspace{2mm}
	\centering
\includegraphics[width=0.22\textwidth]{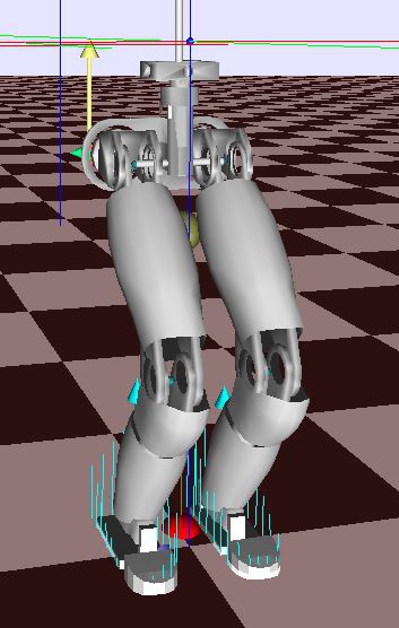}\hfill \includegraphics[width=0.22\textwidth]{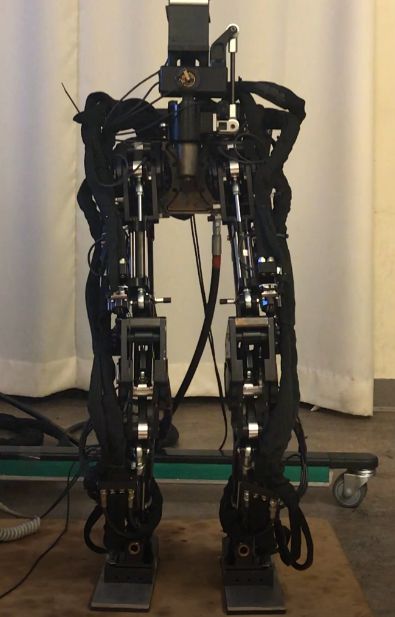}
	\caption{Simulated and real-world lower part of the Sarcos Hermes humanoid robot used in the experiments.}
	\label{fig:front_page}
	\vspace{-4mm}
\end{figure}

\subsection{Problem Statement}
In this paper we investigate task-specific dynamics learning to improve task execution while increasing compliance in the scope of optimization-based inverse dynamics control. Therefore, we pursue the following objectives:
\begin{itemize}
  \item reduce the required contribution of the feedback term in the control,
  \item consequently improve task-space tracking while increasing compliance, and
  \item act directly in the task space of interest, typically the center of mass (CoM) dynamics.
\end{itemize}
Acting directly in the task space as opposed to joint space will potentially
enable applications beyond the scope of the learned dynamics, i.\,e., through generalization.

The intended application of the proposed algorithm is to improve control of dynamic tasks on humanoid robots. We performed our experiments on the lower part of the Sarcos Hermes humanoid robot, depicted in Fig. \ref{fig:front_page}.

This paper is organized as follows:. Section \ref{sec:related_work} provides a short literature overview.  Section \ref{sec:control} gives an overview of the QP inverse dynamics controller used in the paper. Section \ref{sec:task_specific} presents our new algorithm that provides learned torques for task-space inverse dynamics control.
Simulation and real robot experiments are present in Section \ref{sec:results}.
Finally, we discuss our results and conclude in Sections \ref{sec:discussion} and \ref{sec:conclusion}.

\section{RELATED WORK}\label{sec:related_work}
Two bodies of work are relevant to this research: 1) learning and exploiting control torques for improved task execution, and 2) task-space inverse dynamics control of floating-base systems.

Learning control signals to iteratively improve task execution using the aforementioned ILC \cite{Bristow2006}, has been extensively used in robotics \cite{Norrlof2000}. Feedback-error-learning, where the feedback motor command is used as an error signal to train a neural network which then generates a feedforward motor command, was proposed by Kawato \cite{Kawato1990}. The idea was extended for learning of contact interaction. In their work, Pastor et al. \cite{Pastor12} combined learning of forces with dynamical systems to improve force tracking in interaction tasks. Similarly, in \cite{Gams14}, dynamic movement primitives (DMP) \cite{Ijspeert2013} were used in combination with learning of interaction forces.

Learning to improve contact interaction was also applied directly to actuation torques. 
Joint torques along kinematic trajectories were learned and encoded as DMPs in \cite{Steinmetz2015}  and used to increase the accuracy in subsequent executions of in-contact tasks. This approach was also applied to full-sized humanoid robots, for example in \cite{Pollard2002}, where a particular trajectory was run, and the joint torques from that trial were used as the feedforward term on the next trial.
These methods can go beyond mere repetition of the same task. In \cite{Denisa2016}, the authors show how the learned joint-torque signals, encoded in parametric form as compliant movement primitives (CMPs), can be generalized for new task parameters (weight, speed, etc.), and in \cite{Petric_2015} how to effectively learn them. However, because the approach is rooted in joint-torques, generalization is somewhat limited to variations of the same full body motions, which is a strong limitation for highly redundant robots such as humanoids that can perform several concurrent tasks.

Optimization-based inverse dynamics control has become a very popular method to control legged robots \cite{Herzog2016,Stephens2010,escande-ijrr-14} as it allows to directly specify control objectives for multiple tasks, while ensuring priorities between tasks and constraint satisfaction (actuation limits, friction cones, etc). The redundancy of a complex robot such as a humanoid can therefore be optimally exploited.
It is also possible to compute optimal feedback gains in a receding horizon manner directly in task space by leveraging the task reduced dynamics \cite{Kuindersma:2013tz,Herzog2016,Herzog:2015vy}.
Such methods have remarkable capabilities, but are often limited by modeling errors which necessitate to significantly increase feedback gains, which is either very limited by the effective control bandwidth or leads to very stiff behaviors.

Some approaches, as far back as \cite{schaal94}, have proposed to learn task-specific dynamic models that can then be used to synthesize control laws. Iterative methods to compute locally optimal control policies have, for example, recently been used with such learned models \cite{Levine2014}. Iterative repetitions of the process are then used to collect additional data and re-learn a new policy \cite{Deisenroth2011,Levine2014}.
In these approaches, the learned models and resulting control policies operate on the whole robot. It is therefore not clear how other tasks and additional constraints can be incorporated without impairing the resulting behaviors.

In this paper, we learn task-specific feedforward models which take into account the error in dynamic models during task execution and combine them with a Quadratic Programming (QP) based inverse dynamics controller. This allows to significantly improve task tracking while creating more compliant behaviors.

\section{CONTROL}\label{sec:control}
In this Section we  briefly introduce the task-space inverse dynamics controller we use in the paper and that was originally proposed in \cite{Herzog2016}.
We model the floating-base dynamics of a legged humanoid robot as
\begin{equation}
\M(\q)\qdd+\h(\q,\qd)= \Sm^T\tr+\J_c^T\gbf{\lambda}, \label{eq:dynamics}
\end{equation}
with a vector of position and orientation of the robot in space and its joint configurations $\q \in \mathbb{SE}(3) \times \mathbb{R}^n $, the mass-inertia matrix $\M\in \mathbb{R}^{(n+6)\times (n+6)}$, the generalized Coriolis, centrifugal and gravity forces collected in $\h \in \mathbb{R}^{n+6}$, the actuation matrix $\Sm \in \left[\mat{O}_{n \times 6} ~ \mat{I}_{n \times n}  \right]$ and the end-effector contact Jacobian $\J_c \in \mathbb{R}^{6m\times n}$, where $n$ is the number of robot's degrees of freedom, $\tq$ are actuation torques and $\gbf{\lambda}$ are the contact forces.

As discussed in \cite{Herzog2016}, the dynamics can be decomposed into actuated and unactuated (floating base) parts, respectively
\begin{eqnarray}
\M_a(\q)\qdd + \vvc{h}_a(\q,\qd) = \tq +\J_{c,a}^T\gbf{\lambda}, \label{eq:actuated_dyn}\\
\M_u(\q)\qdd + \h_u(\q,\qd) = \J_{c,u}^T\gbf\lambda. \label{eq:un-actuated_dyn}
\end{eqnarray}
and it is only necessary to enforce Eq. \eqref{eq:un-actuated_dyn} to ensure
dynamic consistency as $\tq$ is a redundant variable which can be eliminated
and replaced by a combination of $\qdd$ and $\gbf{\lambda}$ according to Eq. \eqref{eq:actuated_dyn} where necessary. Eq. \eqref{eq:un-actuated_dyn}
is the first constraint to be satisfied by the controller.
Kinematic contact constraints ensuring that the part of the robot in contact with the environment does not move,
\begin{equation}
\J_c \qdd + \dot{\J}_c \qd = 0, \label{eq:constraint2}
\end{equation}
are additional equality constraints for the optimization, where $\J_c$ is the Jacobian for $m_c$ constrained endeffectors. Additionally, we limit foot center of pressure (CoP), friction forces, resultant normal torques, joint torques and joint accelerations with linear inequality constraints. The cost to be minimized is
\begin{equation}
\begin{split}
\min\limits_{\qdd,\lambda}
\sum_{t} ||\xdd_t - \xdd_{t,\mathrm{des}}||^2_{\mat{W}_t} + \\||P_{\mathrm{null}}(\qdd - \qdd_{\mathrm{des}})||^2_{\mat{W}_q} +  ||\lambda - \lambda_{\mathrm{des}}||^2_{\mat{W}_{\lambda}} 
\end{split}
\end{equation}
where $\vvc{x}_t$ are either Cartesian end-effector poses ($\vvc{x}_t \in SE(3))$ or the center of mass position ($\vvc{x}_t \in \mathbb{R}^3$).
The $\mat{W}_t$ are weighting positive definite matrices, $P_{\mathrm{null}}$ projects into the null space of all the Cartesian tasks,  Joint and Cartesian tasks are related through
\begin{equation}
\xdd_t = \J_t\qdd+\dot{\J}_t\qd, \label{eq:cart_joint_accel}
\end{equation}
where $\J_t$ is the Jacobian of the unconstrained end-effector or CoM.
Desired end-effector and CoM accelerations are computed through
\begin{equation}
\xdd_{t,\mathrm{des}} = \xdd_{t,\mathrm{ref}} + \mat{P}_t\left(\vvc{x}_{t,\mathrm{ref}}-\vvc{x}_i\right) + \mat{D}_t\left(\dot{\vvc{x}}_{t,\mathrm{ref}}-\dot{\vvc{x}}_t\right). \label{eq:des_cart_acc}
\end{equation}
Positive definitve matrices $\mat{P}_x$ and $\mat{D}_x$ represent stiffness and damping gains of the PD controller, respectively. Desired joint accelerations are specified by
\begin{equation}
\qdd_\mathrm{des}=\mat{P}_q(\vvc{q}_{\mathrm{ref}}-\q) - \mat{D}_q \dot{\q}. \nonumber
\end{equation}
For more details on the solver, see \cite{Herzog2016}.

\section{Task Specific Dynamics} \label{sec:task_specific}
The controller presented in the previous section allows to track any type of Cartesian task. To improve tracking in task space, we introduce an additional feedforward term.
Indeed, tracking performance depends on the accuracy of the model. This can be seen in (\ref{eq:des_cart_acc}), where
\begin{equation}
 \xdd_{\mathrm{des}} = \underbrace{\xdd_{\mathrm{ref}}}_{\mathrm{feedforward}} +\underbrace{\mat{P}_x\left(\vvc{x}_{\mathrm{ref}}-\vvc{x}\right) + \mat{D}_x\left(\dot{\vvc{x}}_{\mathrm{ref}}-\dot{\vvc{x}}\right)}_{\mathrm{feedback}} .\nonumber
\end{equation}

If the model were perfect, the contribution of the feedback part would amount to 0. However, in the real world it is not, and the feedback part accounts for the discrepancy. We propose recording the feedback contribution part and adding it in the next repetition of the exact same task (desired motion). Thus, we get

\begin{equation}
 \xdd_{\mathrm{des},i} = \underbrace{\xdd_{\mathrm{ref}} + \xdd_{\mathrm{fb},~i-1}}_{\mathrm{updated~feedforward}} +\mat{P}_x\left(\vvc{x}_{\mathrm{ref}}-\vvc{x}\right) + \mat{D}_x\left(\dot{\vvc{x}}_{\mathrm{ref}}-\dot{\vvc{x}}\right), \label{eq:task_specific_dynamics}
\end{equation}
where
\begin{equation}
\xdd_{\mathrm{fb},~i-1} =\xdd_{\mathrm{fb},i-2} + \mat{P}_x\left(\vvc{x}_{\mathrm{ref}}-\vvc{x}_{i-1}\right) + \mat{D}_x\left(\dot{\vvc{x}}_{\mathrm{ref}}-\dot{\vvc{x}}_{i-1}\right). \label{eq:x_fb_i-1}
\end{equation}
It means that at each iteration, we add to the new feedforward term the
previous contributions of the feedback terms, therefore learning the
error dynamics.
The number of learning iterations is arbitrary, depending on the desired accuracy. However, stability of the learning process needs to be ensured, see \cite{Bristow2006} for details on ILC. In our experiments, we only used the feedback from 1 previous iteration (i = 2) as it was sufficient to already significantly improve performance.

Unlike \cite{Pollard2002} or \cite{Denisa2016}, the feedforward part is added in the task-space of the robot, and not in its joint space. It is then combined with the QP-based inverse dynamics controller.

Using the recorded (learned) feedback signal in the next iteration of the same task provides us with an improved feedforward signal, which is task-specific. However, we can build up a database of such signals for different task variations. Thus, the proposed algorithm can significantly correct the discrepancy between the model and the real system. Furthermore, we can use the database to generate an appropriate signal for previously untested tasks and task variations using statistical generalization as in \cite{Denisa2016}.

\subsection{Encoding the Feedforward Signal}
The recorded (learned) signal can be encoded in any form. For example, for end-to-end (discrete) tasks, discrete DMPs can be used \cite{Ijspeert2013}. The use-case example in this paper is periodic squatting. Because our task is periodic, we chose to encode the signal as a linear combination of radial basis functions (RBF) appropriate for periodic tasks. Using RBFs has the advantage that the signal encoding is compact and the signal itself is inherently filtered. As discussed in \cite{Denisa2016}, this representation allows for computationally light\footnote{The calculation of the hyperparameters for GPR is computationally expensive, but it is performed offline. Simple matrix multiplication is performed online.} generalization using Gaussian Process Regression (GPR) \cite{rasmussen2006}. A linear combination of RBF as a function of the phase $\phi$ is given by
\begin{equation}
   \xdd_{\mathrm{fb},~i-1}(\phi) = \frac {\sum_{j=1}^{L} w_{j} \Gamma_j(\phi)} {\sum_{j=1}^{L} \Gamma_j(\phi)},
	\label{eq:basis_functions}
\end{equation}
where $\Gamma_j$ denotes the basis function, given by
\begin{equation}
	 \Gamma_j(\phi) = \exp(h_j(\cos(\phi-c_j)-1)),
	\label{eq:psi}
\end{equation}
$w_j$ is the weight of the $j$-th basis function, $L$ is the number of basis functions, $c_i$ are centers of the basis functions and $h_i>0$ their widths.
The periodic phase $\phi$ is determined by the phase oscillator
\begin{equation}
\dot{\phi}=\Omega,
\label{eq:phi_omega}
\end{equation}
where $\Omega$ is the frequency of oscillations. While in our use-case the phase is linear because of a constant task frequency, it can change over time and even adapt to external signals \cite{Gams2009}.

\subsection{Generalization}
In the manner of generalizing joint-space feedforward torques \cite{Denisa2016}, we can also generalize the learned task-space CoM accelerations. Having chosen RBF encoding, we can generalize between the weights, for example using GPR. The goal of generalization is to provide us with a function
\begin{equation}
\bm{F}_{\bm{D}_b}: \kappa \longmapsto [\vvc{w}]
\label{eq:generalization}
\end{equation}
that provides the output in the form of a vector of RBF weights $w$, given the database of trained feedforward terms $\bm{D}_b$ and the input, i.\,e., the query $\kappa$. We refer the reader to \cite{rasmussen2006,Denisa2016} for details on GPR.
This kind of generalization is analog to the one in \cite{Denisa2016}, but in task space. 

\section{EXPERIMENTAL RESULTS}\label{sec:results}
Experiments were performed on the lower part of the hydraulically actuated, torque controlled, Sarcos humanoid robot. It has in total 17 degrees of freedom (DoFs), with 7 in each leg and three in the torso. The system is depicted in Fig. \ref{fig:front_page}. We used the SL simulation and control environment \cite{Schaal2009} for evaluation in simulation.

To demonstrate the applicability of the approach we use a periodic squatting task. We compare position error of the robot's CoM without and with the added feedforward term, given by (\ref{eq:x_fb_i-1}). Note that any kind of task, be it periodic or an end-to-end task, can be implemented in the same manner. However, for a different kind of task, some other trajectory encoding might be beneficial.

\subsection{Squatting Simulation}
In the first experiment, we performed periodic squatting at a frequency of 0.25~Hz. For the use-case example, squatting was defined as a vertical sinusoidal motion of 3cm amplitude; this range is close to the maximal motion the robot can perform without hitting joint limits. Any other squatting trajectory, obtained for example through motion capture, could also be used.
\begin{figure}[!b]
  \vspace{-4mm}
	\centering
	\includegraphics[width=0.45\textwidth]{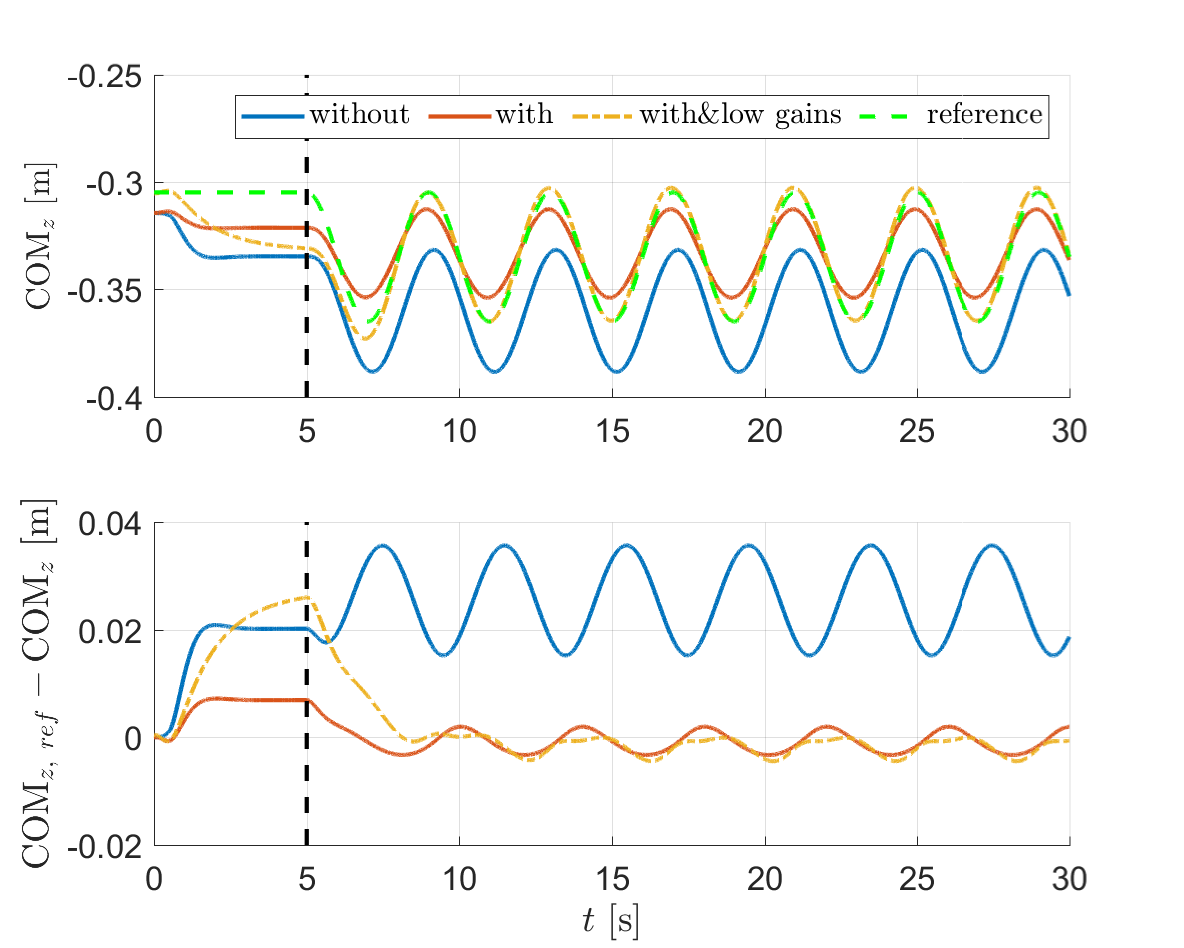}
	\caption{Top: Center of mass position in the vertical $z$ axis during a squatting experiment without the learned feedforward term (blue), with the learned feedforward term (red), and with the feedforward term but with 5 times reduced $\mat{P}$ feedback gains (ocher). Desired CoM motion in dashed-green. Black dashed line depicts the start of squatting. Bottom: Error of CoM$_z$ tracking for all three cases.}\label{fig:cog_err}
\end{figure}

In the simulation experiments, we use a different dynamic model for the inverse
dynamic computation than for the simulation, in order to test the performance
of our approach with modeling errors.
We changed the model so that there was a 10\% error in the mass and inertia matrices of each link. A 20\% difference would not achieve meaningful squatting with the given parameter set; we discuss this in Section \ref{sec:discussion}.

Results in Fig. \ref{fig:cog_err} show the difference in tracking error when using the learned feedforward term, and when not using it. Reduction of tracking error is clearly visible when the learned feedforward term is used. Furthermore, when using the learned feedforward term, we can significantly reduce the $\mat{P}$ gain (in this case 5X) without visibly increasing the tracking error. This demonstrates that the contribution of the feedback term is low and that increased compliance can be achieved. Before the beginning of squatting (depicted by the black dashed line), the learned feedforward term was a constant value learned for the start of the squat from steady-state squatting. This does not completely match the initial posture of the robot in simulation, which has a small starting randomness built-in. The higher error in the starting posture when using low gains indicates higher compliance of the robot (i.e. the errors of the model induce an error in steady state positions).

\begin{figure}[b!]
	\centering
	\includegraphics[width=0.45\textwidth]{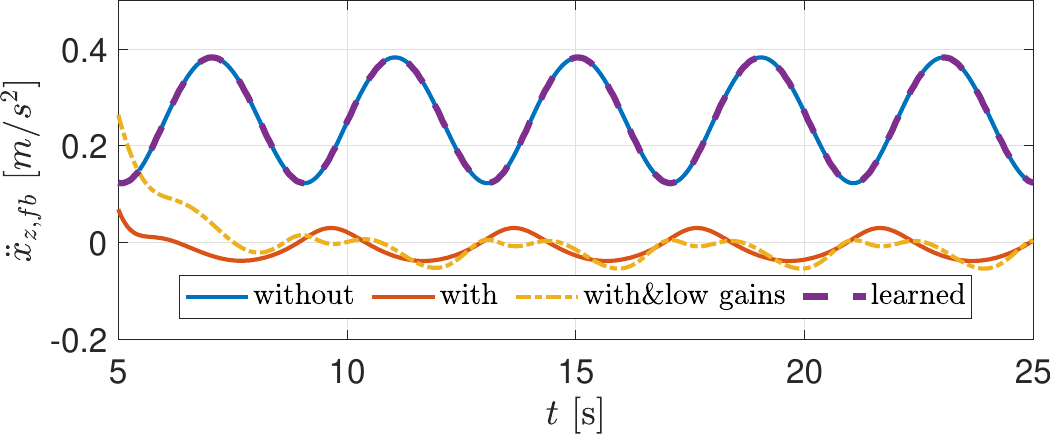}
	\caption{The value of the feedback term when squatting as defined in the experiment. Without the additional feedforward term in blue, with the additional term in red, with the term but with reduced gains in ocher, and the encoded feedforward signal in dashed purple.}
	\label{fig:RBS_error}
\end{figure}

The plot in Fig. \ref{fig:RBS_error} shows the amplitude of the feedback part of the controller for the same three cases as for Fig. \ref{fig:cog_err}. We can again see the difference in the necessary feedback correction. Feedback correction is by an order of magnitude larger if the learned feedforward term is not used. The plot also shows that the encoded feedforward signal (purple) very closely matches the original feedback signal. The matching could be increased, for example, with a higher number of basis functions. In the experiments we used $L = 25$ basis functions; the number was chosen empirically. 

\subsection{Generalization to different squatting amplitudes}
\begin{figure}[!t]
	\vspace{2mm}
	\centering
	\includegraphics[width=0.5\textwidth]{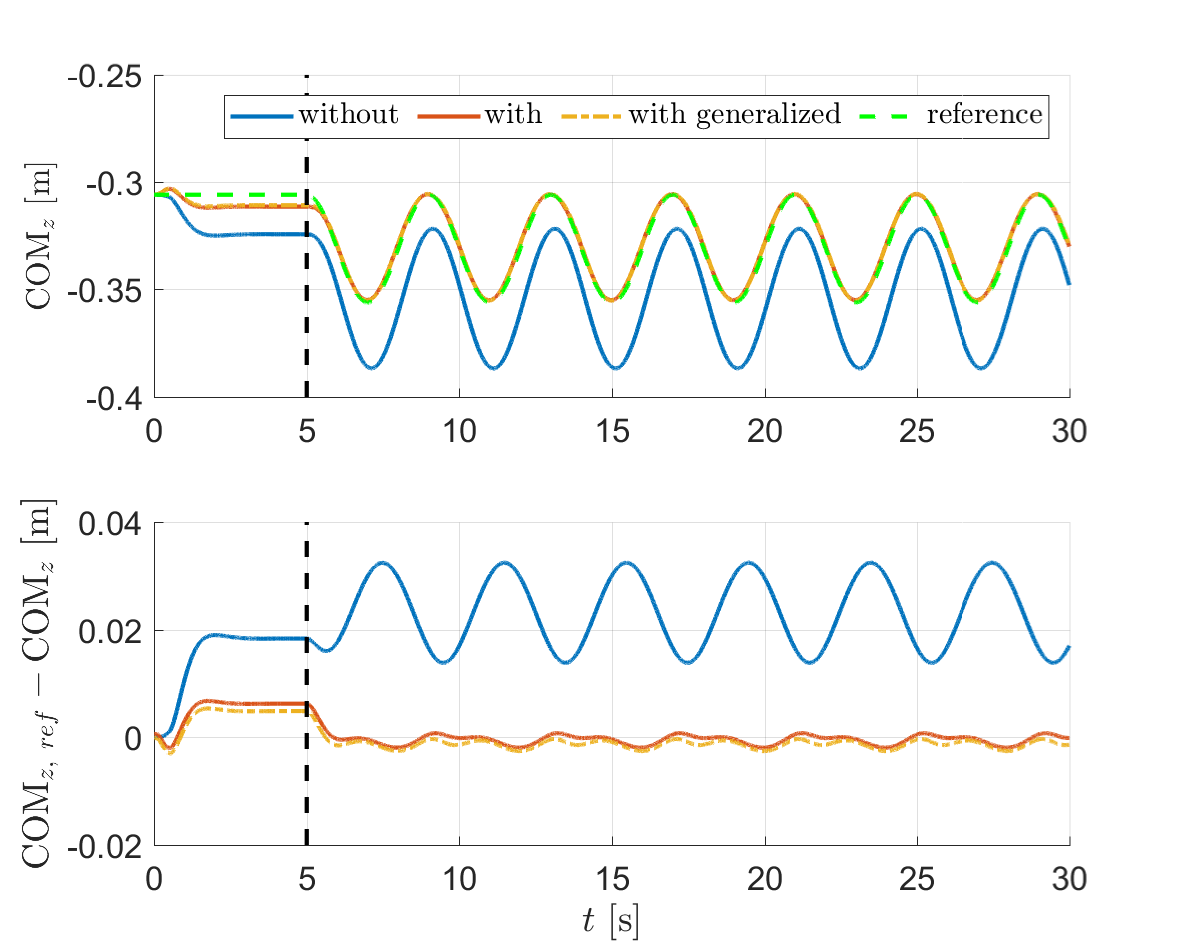}
	\caption{Top: Center of mass position in the vertical $z$ axis during a squatting experiment without the learned feedforward term (blue), with the learned feedforward term (red), and with the feedforward term generalized from a database (ocher). Desired CoM motion in dashed-green. Black dashed line depicts the start of squatting. Bottom: Error of CoM$_z$ tracking for all three cases.}\label{fig:gen_cog_err}
	\vspace{-2mm}
\end{figure}

We performed a generalization experiment over a variation of the task, to show that the approach can be used with generated feedforward signals. We used GPR to generalize the feedforward term for a squatting amplitude of $\kappa=5$~cm. The database consisted of feedforward terms for different squatting amplitudes\footnote{The database is too small. Typically it would consist of tens of entries, but in this toy example they would be too close together.} $\kappa={2,4,6,8}$~cm. Fig. \ref{fig:gen_cog_err} shows that the generalized feedforward term allowed for very similar, low CoM$_z$ tracking errors as the recorded feedforward term for $\kappa=5$~cm. While this kind of generalization is similar to the one represented in \cite{Denisa2016}, in our case the generalization was in task space, i.\,e., generalization was between the weights for 1 DoF. To achieve the same in joint space would require generalization for all 17 DoF of the robot. Our approach therefore leads to a simpler generalization function
thanks to the combination of the task-space feedforward term and the inverse dynamics controller.

\subsection{Real Robot Experiments}
\begin{figure}[!b]
  \vspace{-4mm}
	\centering
	\includegraphics[width=0.45\textwidth]{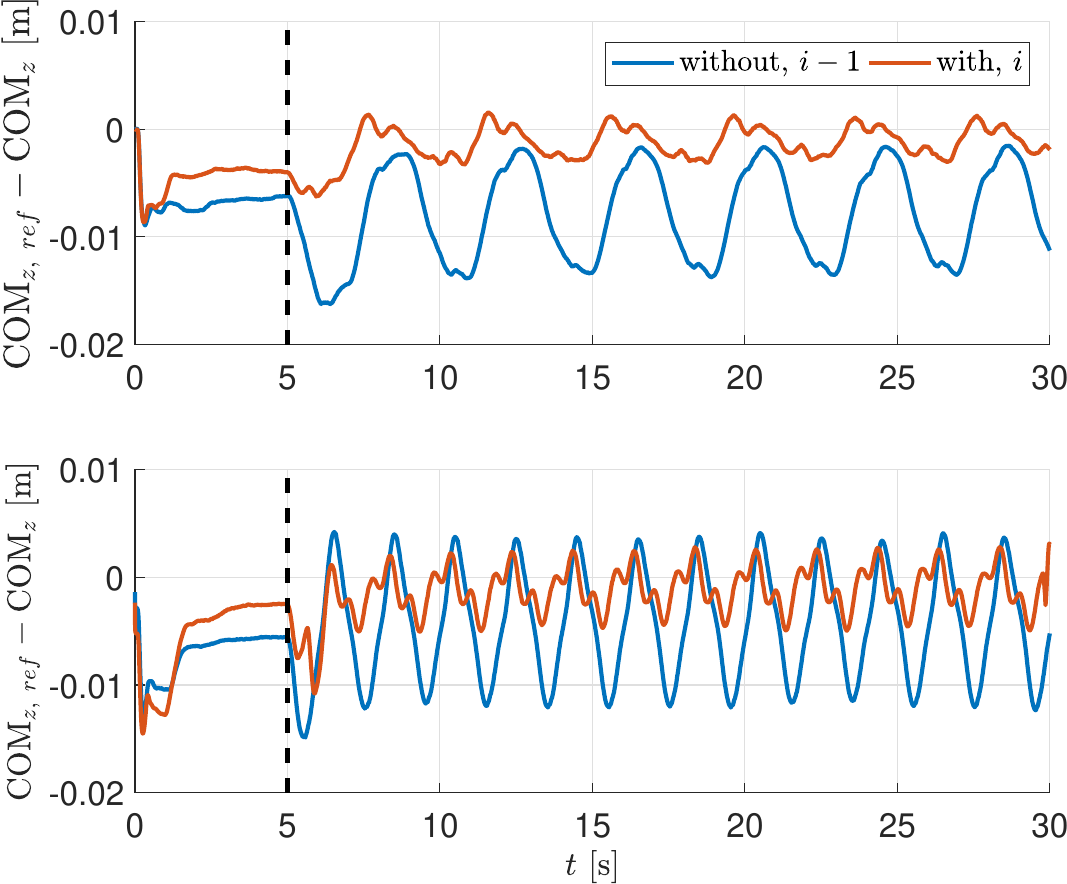}
	\caption{Real-world error of CoM$_z$ tracking with and without the added term for the squatting experiment at two different squatting frequencies, 0.25~Hz in the top and 0.5~Hz in the bottom. In both plots the results without the added term are in blue, and with the added feedforward term in red.}\label{fig:realcog_err}
	\vspace{-2mm}
\end{figure}

\begin{figure*}
\vspace{2mm}
\includegraphics[width=0.1\textwidth]{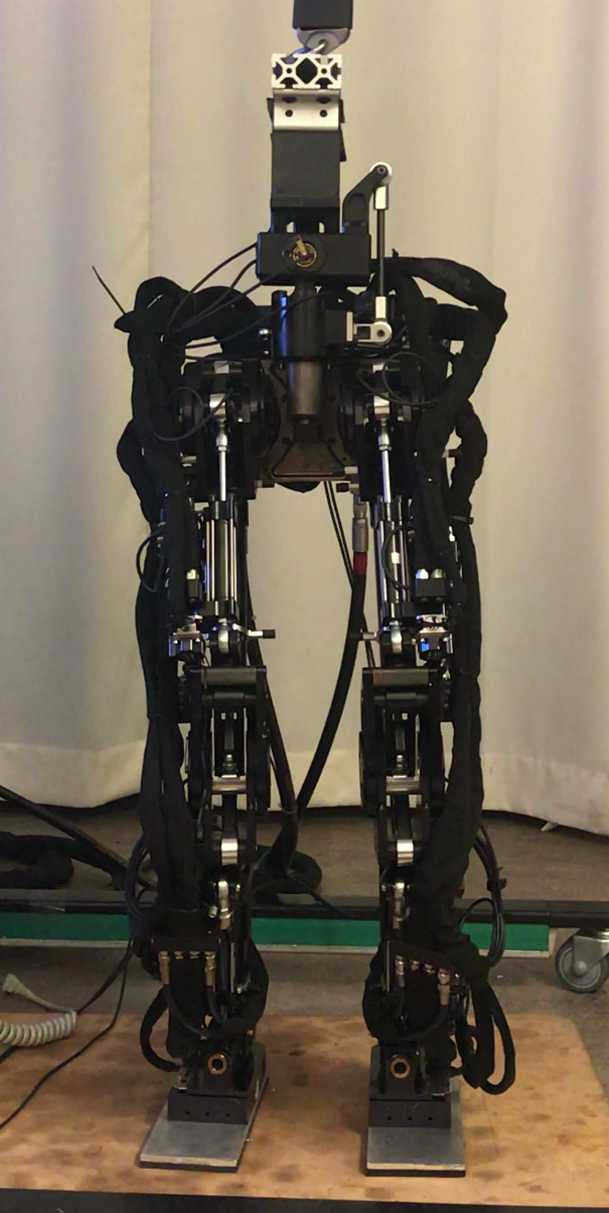}  \hfill
\includegraphics[width=0.1\textwidth]{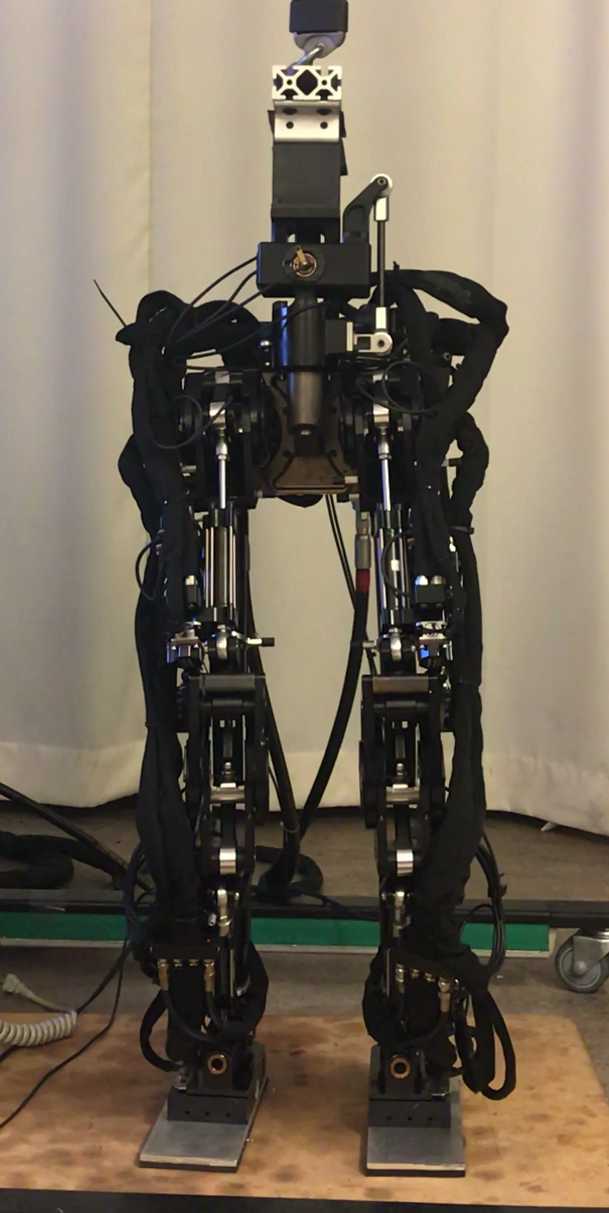}  \hfill
\includegraphics[width=0.1\textwidth]{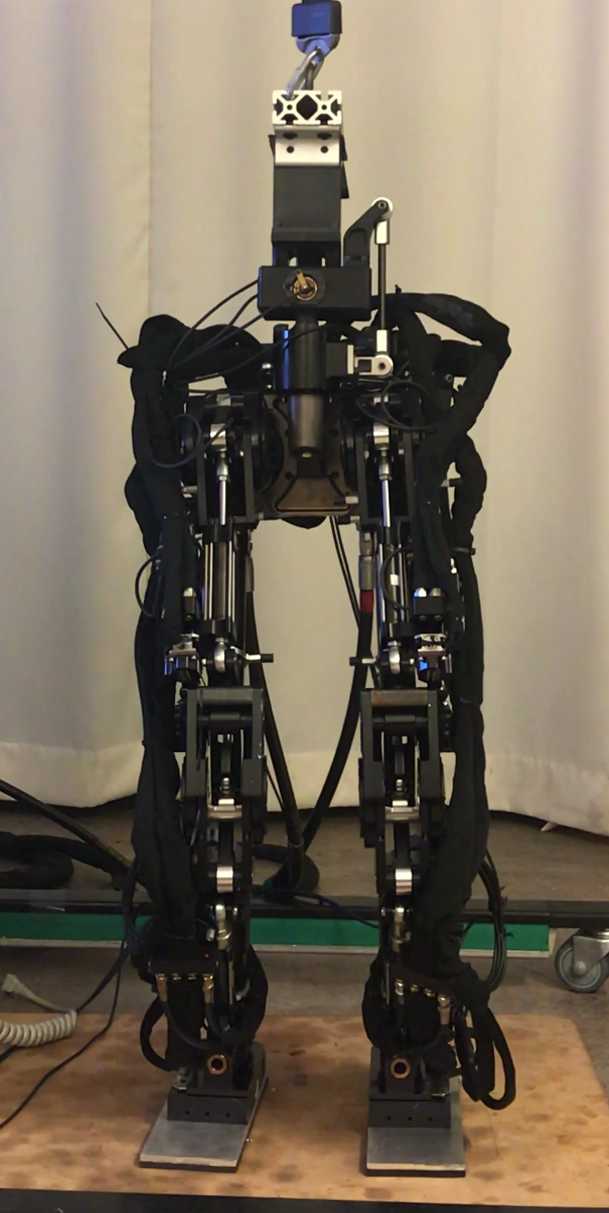}  \hfill
\includegraphics[width=0.1\textwidth]{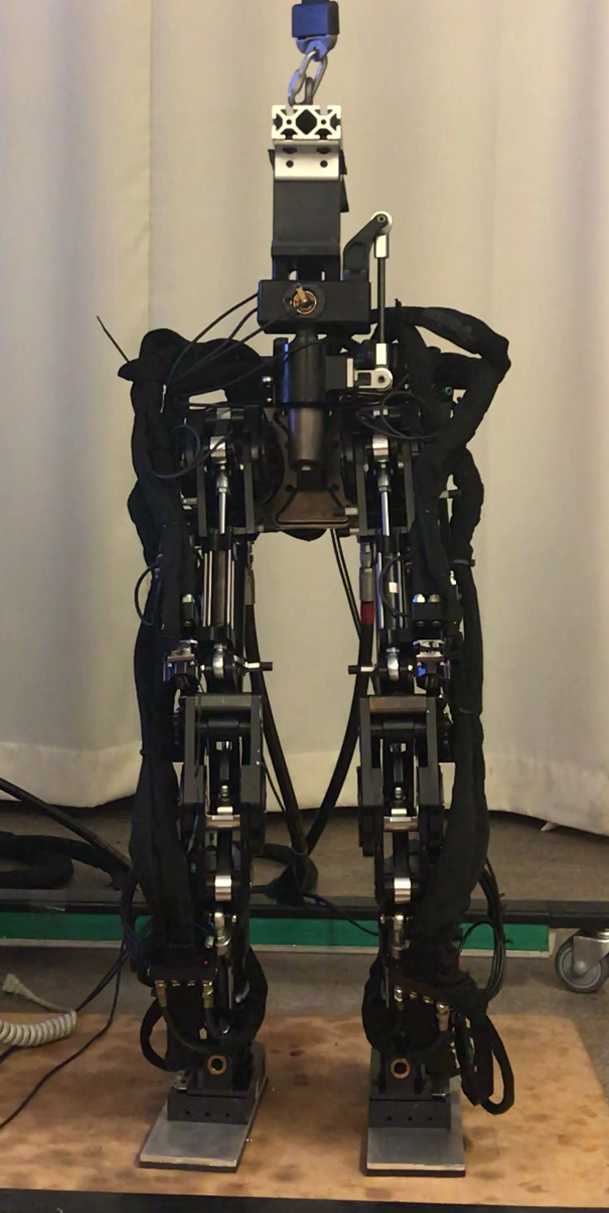}  \hfill
\includegraphics[width=0.1\textwidth]{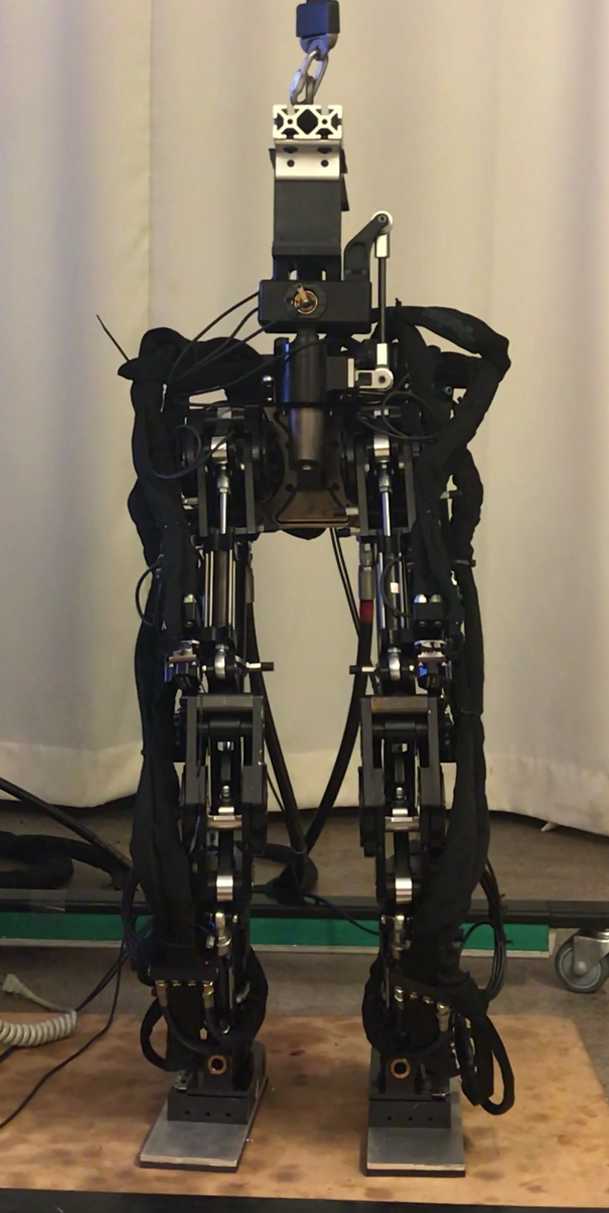}  \hfill
\includegraphics[width=0.1\textwidth]{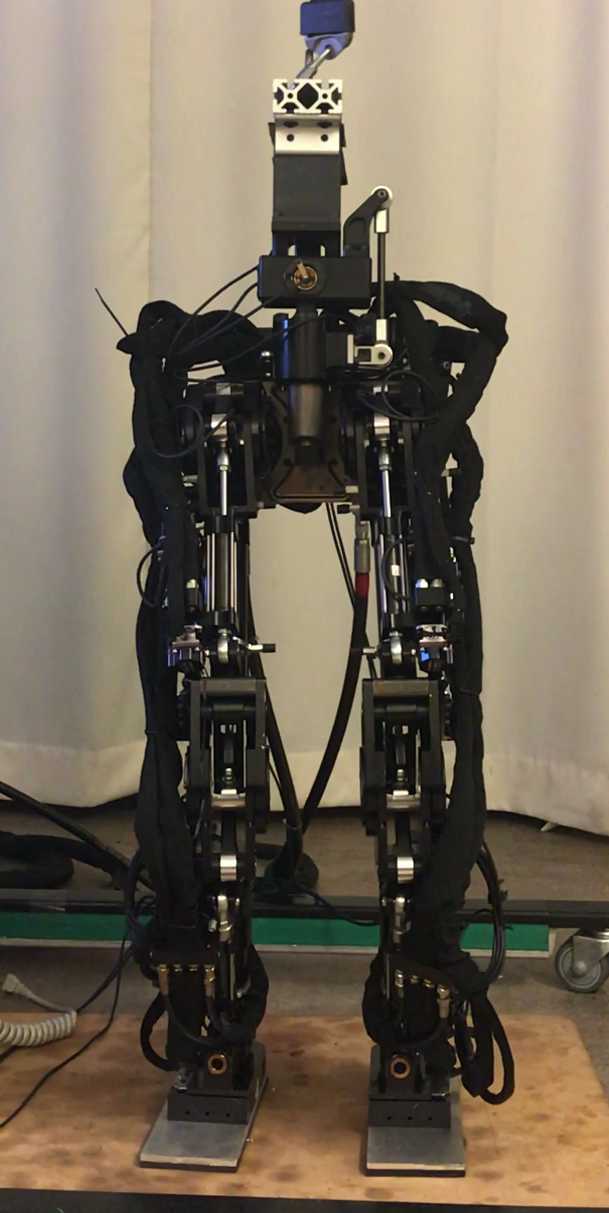}  \hfill
\includegraphics[width=0.1\textwidth]{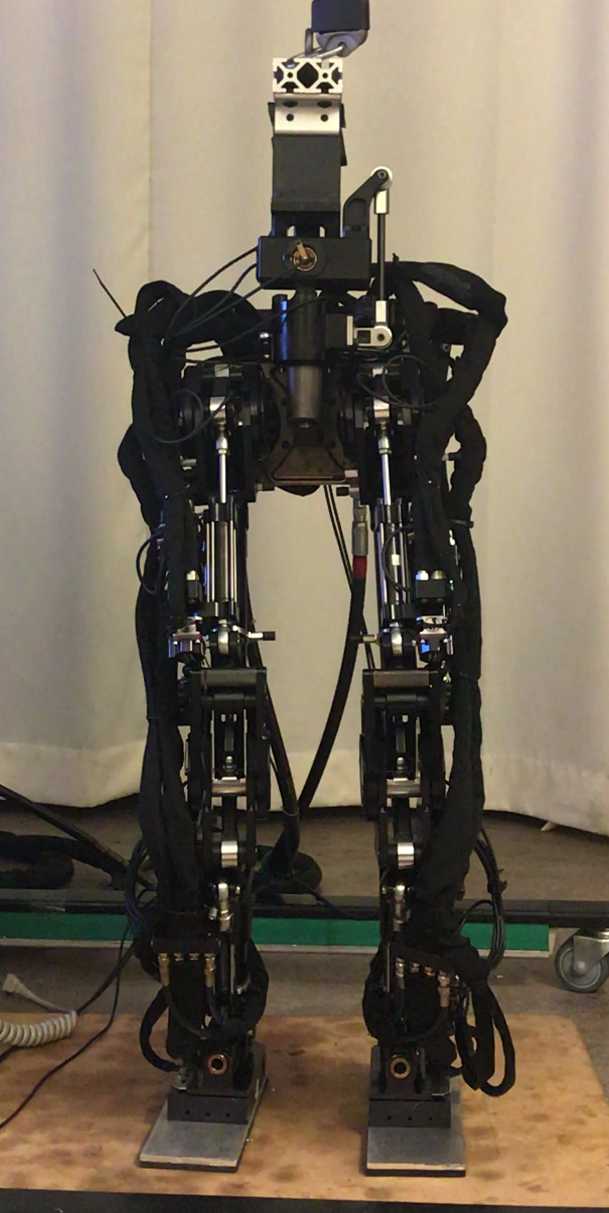}  \hfill
\includegraphics[width=0.1\textwidth]{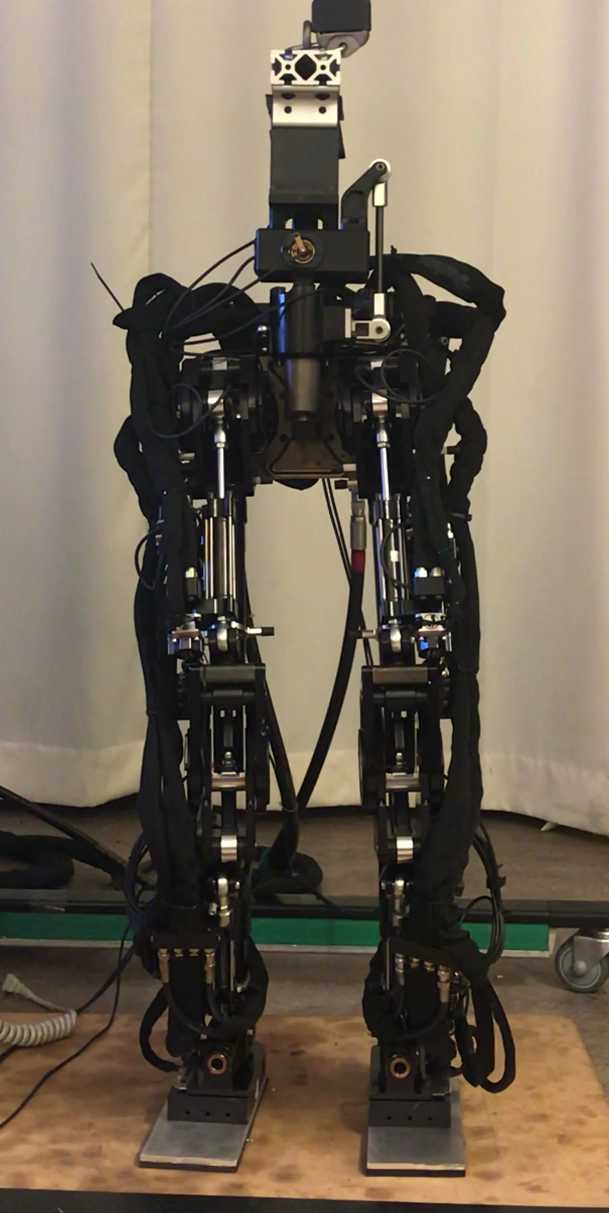}  \\
\vspace{-6mm}
\caption{Still images of the real-world lower part of the Sarcos Hermes robot performing one squat. 
See also the accompanying video.} \label{fig:sequence}
\vspace{-5mm}
\end{figure*}

We performed the same experiment on the real robot. To vary the dynamics of the task, we tested our approach for two different squatting frequencies: 0.25~Hz and 0.5~Hz. The error of CoM$_z$ tracking for both cases is depicted in Fig. \ref{fig:realcog_err}.
We can see in the plots that the error is again significantly reduced for both cases. Small oscillations in the behavior are the consequence of acting on a real system with an imperfect model and feedback signals. CoM position on the real system was estimated using the joint encoders and the kinematics, with assumed flat feet on the ground.
Fig. \ref{fig:sequence} shows a series of still photos depicting the real system performing one squat.

\section{DISCUSSION}\label{sec:discussion}
We have proposed a method to learn task-space dynamics which allows to
improve task performance while reducing feedback gains.
Both the simulation and real robot results show a reduced contribution of the feedback term; a clear improvement of the tracking performance and the possibility to reduce the feedback gains to thus increase the compliance. In the following, we briefly discuss these points.

We first discuss the improvement of the tracking performance and the reduction of the feedback term contribution. The question whether the system behavior is the same with the additional task-specific feedforward term and low feedback gains or without the additional task-specific feedforward term and high gains has previously been studied \cite{GOLDSMITH2002,Verwoerd_2002}. An equivalent feedback can always be constructed from the ILC parameters with no additional plant knowledge for proper causal LTI systems, whether or not the ILC includes current-cycle feedback. Our system (\ref{eq:task_specific_dynamics}) is not proper causal, therefore complete equivalence cannot be claimed, but the results show a clearly reduced contribution of the feedback term. Herein lies the main advantage of using the proposed approach -- lower feedback gains can be used, resulting in increased compliance. Compliance of the system has been recognized as one of the key elements for real-world deployment of robots in unstructured environments, as it provides robustness for unplanned disturbances \cite{Stephens2010}.

Our approach reduces the contribution of the feedback term in a manner similar to an improved dynamic model. As shown in the literature (e.\,g. in \cite{Mistry2009}), a dynamic model never completely describes the behavior of a complex system. On a real system, such as the lower part of the Hermes humanoid robot used in this work,
unmodeled hydraulic hoses, actuator dynamics, friction and flexibilities can have
a significant effect.
The originality of our approach is that we improve a task-specific dynamic model.
The learned feedback torques for squatting are by default only applicable to squatting. As already outlined in Section \ref{sec:control}, building up a database is a rather straightforward solution. This has not only been applied to the \emph{model} (for lack of a better word), but also to control policies as a result of optimization. In \cite{Mansard2018}, a database of such control policies is used to warm-start the optimization. A more advanced solution than just building up a database is to use the database to generate feedforward terms for previously untrained situations. Different methods can be applied, for example statistical learning, such as GPR, which was used in a similar manner for joint torques in \cite{Denisa2016}. 

As shown in Section \ref{sec:results}, one of the advantages of the proposed approach of generalization in the task space is the reduced dimensionality of the task. Another advantage of learning and applying the feedforward terms in task space is the similarity it has over different tasks. The proposed approach requires first a working solution, so that the feedforward term can be learned. However, this working solution might be difficult to achieve for complex tasks if the model is not sufficiently accurate. Even our squatting use-case will not work if the model is 20\% off. Achieving a complex task can be challenging but we posit that the feedforward term of a less-complex task could be used to bootstrap the execution of the more complex task. During walking, the CoM position is moving from one side to the other, which is (from the CoM point of view) the same as simply shifting the weight from one side to the other without lifting the feet. 
With feedforward terms for shifting of the weight from one side to the other, which makes the execution of this task more accurate, it is only one (algorithmic) step to implement stepping in place.
This cross-task generalization, however, remains an open research question. Furthermore, such generalization can only be applied over similar tasks, and steps that ensure that the feedforward term does not worsen the solution need to be taken.

\section{CONCLUSION \& FUTURE WORK} \label{sec:conclusion}
In this paper we showed that basic iterative learning can be applied to task-space accelerations in order to improve the task execution of a complex, free-floating robot system, controlled with task-space inverse dynamic controllers and that generalization to variations of the task is also possible. Furthermore, it has the potential to improve the behavior of model-based control methods with the application of generalized signals for different task parameters, and possibly even across different task. The latter, however, remains for future work.




\bibliographystyle{ieeetr}
\bibliography{IEEEabrv,bibIROS18}

\end{document}